\documentclass{article}
\usepackage{spconf,amsmath,graphicx,hyperref}
\usepackage{graphicx}
\usepackage{amsmath}
\usepackage{amssymb}
\usepackage{makecell}
\usepackage{booktabs}
\usepackage{algorithm}
\usepackage{xcolor}
\usepackage{caption}
\usepackage{subcaption}
\usepackage{enumitem}

\DeclareMathOperator*{\argmin}{arg\,min}
\usepackage{algorithmic}
\title{Deep Image Prior with L0 Gradient Regularizer for Image Smoothing}
\name{Nhat Thanh Tran, Kevin Bui, Jack Xin\thanks{The work was partially supported by NSF grants DMS-2151235, DMS-2219904, DMS-2309520, and a Qualcomm Gift Award. NTT was also funded by a Faculty Endowed Fellowship and the Graduate Scholar Success Fund from the University of California, Irvine.}}
\address{Department of Mathematics; University of California, Irvine; Irvine, CA 92697, United States}
\begin{document}
\ninept
\maketitle
\begin{abstract}
Image smoothing is a fundamental image processing operation that preserves the underlying structure, such as strong edges and contours, and removes minor details and textures in an image. Many image smoothing algorithms rely on computing local window statistics or solving an optimization problem. Recent state-of-the-art methods leverage deep learning, but they require a carefully curated training dataset. Because constructing a proper training dataset for image smoothing is challenging, we propose DIP-$\ell_0$, a deep image prior framework that incorporates the $\ell_0$ gradient regularizer. This framework can perform high-quality image smoothing without any training data. To properly minimize the associated loss function that has the nonconvex, nonsmooth $\ell_0$ ``norm", we develop an alternating direction method of multipliers algorithm that utilizes an off-the-shelf $\ell_0$ gradient minimization solver. Numerical experiments demonstrate that the proposed DIP-$\ell_0$ outperforms many image smoothing algorithms in edge-preserving image smoothing and JPEG artifact removal.
\end{abstract}

\section{Introduction}
As one of the most fundamental operations in computer vision and image processing, image smoothing decomposes an image into the base layer and the detail layer. Being piecewise smooth, the base layer contains the image's structural information, such as edges and contours; being sparse, the detail layer contains the image's residual information, such as details and textures. This image decomposition has several applications, including clip-art compression artifacts removal \cite{nguyen2015fast, xu2011image} and texture removal \cite{cho2014bilateral,xu2012structure}.

Image smoothing algorithms are classified into three main groups: local filters, global filters, and deep filters. Local filters compute the weighted average of a pixel and its associated neighbors within a window to smooth the input image. Examples include bilateral filter \cite{tomasi1998bilateral} and guided image filter \cite{he2012guided}. Although local filters are computationally efficient, they are prone to creating artifacts, such as halos and gradient reversals, because they rely on computing local image statistics. Global filters, on the other hand, solve an optimization problem whose objective function typically consists of a fidelity term for approximating the original image and a regularizer for smoothing the original image. Global filters include weighted least square (WLS) \cite{farbman2008edge, liu2024fast}, $\ell_0$ smoothing \cite{nguyen2015fast, xu2011image,pan20240}, relative total variation \cite{xu2012structure, li2023enhance}, and bilateral total variation \cite{he2022structure}. Unlike local filters, global filters avoid creating halos and gradient reversals, but they are computationally expensive. Lastly, deep filters are deep neural networks (DNNs) that learn to emulate either local or global filters. They include deep edge-aware filters \cite{xu2015} and JESS-Net \cite{feng2022}. Although deep filters demonstrate outstanding performance, they need to be trained on a large dataset consisting of the original, natural images as inputs and their smoothed counterparts as targets so that they can be applicable for any kind of images. Typically, the smoothed counterparts are created by other smoothing methods, thereby limiting deep filters' potential to generate better smoothing results. As of now, the NKS dataset \cite{xu2021} and the SPS dataset \cite{feng2022} are the only datasets that have the smoothed images as the original, natural images and the ``noisy" images generated from the smoothed images via texture blending. Hence, training datasets for supervised deep filters are currently limited.

To circumvent the lack of high-quality training data for image smoothing, we propose an unsupervised  deep filter based on deep image prior (DIP) \cite{ulyanov2018deep}. DIP has demonstrated remarkable performance without any training data in image denoising, image inpainting, and image superresolution by solving them as optimization problems. Since image smoothing can be formulated as an optimization problem as done by global filters, we demonstrate that it can be achieved by DIP when incorporating the $\ell_0$ gradient regularizer. The $\ell_0$ gradient regularizer preserves strong edges and boundaries in images, but any general $\ell_0$ minimization problem is difficult to solve because of its nonconvexity and nonsmoothness. Therefore, this work not only proposes a DIP framework with $\ell_0$ gradient regularizer specifically for image smoothing but also designs a specialized optimization algorithm for it.

The main contributions of our paper are summarized as follows. First, we propose a DIP framework for image smoothing by incorporating the $\ell_0$ gradient regularizer. As a result, it is an unsupervised deep filter that resembles a global filter because it solves an optimization problem. Second, instead of using the subgradient method as done in \cite{ulyanov2018deep,liu2019image}, we minimize the loss function of DIP with alternating direction method of multipliers (ADMM) \cite{boyd2011distributed} that utilizes an approximation algorithm proposed in \cite{nguyen2015fast}   for $\ell_0$ gradient minimization. Third, in a series of experiments, we demonstrate that our proposed method performs well against other state-of-the-art algorithms in image smoothing and JPEG artifact removal.
\section{Review of Deep Image Prior}
Utilizing the superior performance of DNNs without any training data for solving classical imaging problems such as image denoising, DIP \cite{ulyanov2018deep} solves the optimization problem
\begin{align}\label{eq:DIP}
\min_{\theta} \|f-g_{\theta} (x)\|_2^2,
\end{align}
where $f$ is the input image, $x$ is a fixed but random input, and $g_{\theta}(\cdot)$ is a DNN parameterized by the weights $\theta$. With $\theta^*$ as the optimal solution to \eqref{eq:DIP}, the output of the DNN $g_{\theta^*}(x)$ is the restored image for the given image processing task. \eqref{eq:DIP} can be further improved by adding an explicit regularizer:
\begin{align}\label{eq:DIP_R}
    \min_{\theta} \|f-g_{\theta} (x)\|_2^2 + \lambda \mathcal{R}(g_{\theta}(x)).
\end{align}
Using total variation (TV) \cite{rudin1992nonlinear} as a regularizer improves DIP's performance \cite{liu2019image}. Usually, \eqref{eq:DIP} and \eqref{eq:DIP_R} are solved entirely by stochastic gradient descent (SGD) via backpropagation \cite{ulyanov2018deep, liu2019image}. However, when the regularizer $\mathcal{R}(\cdot)$ is non-smooth, \eqref{eq:DIP_R} is solved by ADMM, as done in \cite{cascarano2021combining}. To our knowledge, DIP has not yet been applied to image smoothing. Therefore, in this paper, we demonstrate its extension to image smoothing by specifying the appropriate regularizer in \eqref{eq:DIP_R} and tailoring an ADMM algorithm for it.

\section{Proposed Method}
Setting  $\mathcal{R}(\cdot)$ in $\eqref{eq:DIP_R}$ as the $\ell_0$ gradient regularizer, we propose DIP-$\ell_0$ for image smoothing:
\begin{align} \label{eq:dip_l0}
    \min_{\theta} \|f-g_{\theta} (x)\|_2^2 + \lambda \|\nabla g_{\theta}(x)\|_0.
\end{align}
The $\ell_0$ gradient regularizer $\|\nabla u\|_0$ counts the number of non-zero gradients in an image  $u$ \cite{xu2011image}. It enforces the output image $g_{\theta}(x)$ to be piecewise constant, which helps preserve corners and sharp edges.

\begin{algorithm}[t!]
\scriptsize
\caption{ADMM Algorithm for DIP-$\ell_0$}
\label{alg:ADMM_DIP}
\begin{algorithmic}[1]
\REQUIRE \makebox[0pt][l]{\begin{minipage}[t]{\linewidth}

\begin{itemize}[noitemsep, topsep=0pt]
    \item $f \in [0,1]^3$ - normalized input image
    \item $g_{\theta}(x)$ - neural network architecture with\\ fixed input $x$
    \item $\lambda > 0$ - regularizer parameter
    \item $\beta > 0$ - penalty parameter
    \item $T$ - number of ADMM iterations
    \item $K$ - number of SGD iterations for solving \eqref{eq:theta_sub}
    \item $\alpha$ - learning rate for SGD
    \item $\gamma$ - weighing parameter for \eqref{eq:exp_sol}
\end{itemize}
\end{minipage}}
\ENSURE Smoothed image $u^T$
\STATE Initialize $v^0, w^0$ as random uniform [0,1] tensors with the same dimension as the input image $f$
\STATE Initialize $u^0 = f$
\STATE Initialize $\theta_K^0$ randomly as weights for the neural network $g_{\theta}$
\FOR{$t \gets 0$ to $T-1$}
    \STATE Set $\theta_0^{t+1} = \theta_K^{t}$
    \FOR {$k \gets 0$ to $K-1$}
    \STATE Update $\theta^{t+1}_{k+1}$ by SGD or variant with learning rate $\alpha$ using backpropagation
    \ENDFOR
    \STATE Update $v^{t+1}$ by solving Region Fusion 
    \cite{nguyen2015fast}
    \STATE $w^{t+1} = w^t + \beta(v^{t+1} - g_{\theta^{t+1}_K}(x))$
    \STATE $u^{t+1} = \gamma g_{\theta_K^{t+1}}(x)+ (1-\gamma)u^t$
\ENDFOR{}
\end{algorithmic}
\end{algorithm}
 
 We assume that the fidelity term $\mathcal{L}(f, g_{\theta}(x)) = \|f-g_{\theta} (x)\|_2^2$ is differentiable with respect to the weights $\theta$. We cannot solve \eqref{eq:dip_l0} directly by SGD via backpropagation because the subdifferential of \eqref{eq:dip_l0} with respect to the weights $\theta$ \cite[Exercise 8.8]{rockafellar2009variational} is $ \nabla_{\theta} \mathcal{L}(f, g_{\theta}(x)) + \lambda \partial_{\theta} \left(\|\nabla g_{\theta}(x)\|_0 \right)$.  By \cite[Theorem 1]{le2013generalized}, the subdifferential of the $\ell_0$ ``norm" is 
 \begin{align*}
     \partial\|x\|_0 = \left\{y \in \mathbb{R}^n: y_i = 0 \text{ if } x_i \neq 0 \right\} \; \text{ for } x \in \mathbb{R}^n. 
 \end{align*}
 This implies that the subdifferential $\partial_{\theta} \left(\|\nabla g_{\theta}(x)\|_0 \right)$ can have up to infinitely many subgradients whenever $(\nabla g_{\theta}(x))_i = 0$ at some pixel $i$. Because $\nabla g_{\theta} (x)$ should be mostly nonzero at the beginning of training, any subgradient of $\partial_{\theta} \left(\|\nabla g_{\theta}(x)\|_0 \right)$ would have mostly zero entries. Therefore, the subgradient of the $\ell_0$ norm is  ambiguous to define for backpropagation and it would barely update the weights $\theta$ to obtain a smoothed image output. Hence, we propose an ADMM algorithm to effectively solve \eqref{eq:dip_l0}.

We reformulate \eqref{eq:dip_l0} by introducing an auxiliary variable $v = g_\theta(x)$, giving us the following constrained optimization problem:
\begin{equation}\label{eq:prob_form}
\begin{aligned}
   \min_{\theta, v}\quad &\|f-g_{\theta} (x)\|_2^2 + \lambda \|\nabla v\|_0 
   \quad \text{s.t.}\quad v =  g_\theta(x).
\end{aligned}
\end{equation}
Thus, the augmented Lagrangian is given as follows:

\begin{equation}\label{eq:aug_Lagrangian}
\begin{aligned}  \|f - g_{\theta} (x)\|_2^2 + \lambda \|\nabla v\|_0 +\langle w,  v -g_{\theta}(x)  \rangle   + \frac{\beta}{2} \|v - g_{\theta}(x) \|_2^2,
\end{aligned}
\end{equation}
where $\beta>0$ is the penalty parameter and $w$ is the Lagrange multiplier. The ADMM algorithm consists of sequentially updating the three unknown variables $\theta, v, $ and $w$ in  \eqref{eq:aug_Lagrangian}. For each iteration $t$, we update as follows:

\begin{subequations}
\begin{align}
     \theta^{t+1} \in \argmin_{\theta} &  \,\|f - g_{\theta} (x)\|_2^2 + \frac{\beta}{2} \left \|v^t - g_{\theta}(x) + \frac{w^t}{\beta} \right \|_2^2  \label{eq:theta_sub},   \\
     v^{t+1} \in \argmin_v & \, \dfrac{2\lambda}{\beta} \|\nabla v\|_0 + \left \|v -  \left(g_{\theta^{t+1}}(x)  - \frac{w^t}{\beta}\right ) \right \|_2^2,  \label{eq:v_sub}\\
     w^{t+1} = w^t + &\beta(v^{t+1} - g_{\theta^{t+1}}(x)).\label{eq:w_sub}
\end{align}
\end{subequations}
\eqref{eq:theta_sub} is solved inexactly by applying SGD using backpropagation for a fixed number of iterations.
Although \eqref{eq:v_sub} is challenging to solve, there are approximation strategies available \cite{nguyen2015fast, xu2011image, storath2014fast}. In our implementation, we choose  Region Fusion \cite{nguyen2015fast} for its computational efficiency and accuracy. It solves \eqref{eq:v_sub} by evaluating whether to fuse neighboring regions of similar pixel intensities into one region of the same intensity. Following \cite{ulyanov2018deep}, we apply exponential averaging to the output $g_{\theta^t}(x)$ as follows:
\begin{align}\label{eq:exp_sol}
    u^{t+1} = \gamma g_{\theta^{t+1}}(x)+ (1-\gamma)u^t,
\end{align}
where $\gamma \in (0,1]$ is a weighing parameter and $u^0 = f$. The final output for \eqref{eq:dip_l0} is $u^{T}$ when we run the ADMM algorithm for $T$ iterations. Algorithm \ref{alg:ADMM_DIP} summarizes the overall ADMM algorithm for \eqref{eq:dip_l0}.

\section{Numerical Experiments}

We evaluate the performance of our proposed model DIP-$\ell_0$ in edge-preserving image smoothing and JPEG artifact removal. By having \eqref{eq:v_sub} solved by Region Fusion \cite{nguyen2015fast}, we have Region Fusion DIP-$\ell_0$, where the source code is provided at \url{https://github.com/kbui1993/Official_L0_Gradient_DIP}. We compare our methods to fourteen other image smoothing algorithms: PNLS \cite{xu2021}, $\ell_0$ smoothing \cite{xu2011image}, Pottslab \cite{storath2014fast}, Region Fusion \cite{nguyen2015fast}, GSF \cite{liu2021generalized}, RTV \cite{xu2012structure}, semi-global WLS \cite{liu2017semi}, semi-sparsity filter \cite{huang2023semi}, JESS-Net \cite{feng2022}, ResNet \cite{zhu2019benchmark}, VDCNN \cite{zhu2019benchmark}, deep WLS \cite{yang2024weighted}, DIP \cite{ulyanov2018deep}, and DIP-TV \cite{liu2019image}. The source codes are obtained from the authors' websites. For only the deep filters JESS-Net, ResNet, VDCNN, and deep WLS, we use the available pre-trained models, while for the others, we tune their parameters. 

To ease parameter tuning, we normalize any input image $f$ to have their pixel intensities in $[0,1]$. The neural network $g_{\theta}$ that we use for all our experiments is the default network used for DIP \cite{ulyanov2018deep}. It is an encoder-decoder with skip connections between its up and down layers. The input $x$ of the neural network $g_{\theta}$ is a random 32-channel image that has the same height and width as the input image $f$. For Algorithm \ref{alg:ADMM_DIP}, we tune $\lambda \in \{0.025, 0.05, 0.075\}$ and $\beta \in \{1.5, 1.75, 2.0, 2.25\}$. To solve \eqref{eq:theta_sub},  we use ADAM \cite{kingma2014adam} with learning rate $\alpha \in \{10^{-3}, 10^{-4}, 10^{-5}\}$ for $K=25$ iterations. For exponential averaging, we set $\gamma = 0.9$. Outputs of Algorithm \ref{alg:ADMM_DIP} are examined at $T \in \{100, 200, 300\}$. For each image processing task, we select the parameter combination that provides the best results on average. 
\subsection{Edge-Preserving Image Smoothing}
\begin{table}[!t]
	\centering
    \scriptsize
    	\caption{Average PSNR and SSIM on 120 randomly selected images from NKS dataset. \textbf{Black bold} is best; \textcolor{blue}{\textbf{blue bold}} is second best.}
	\begin{tabular}{|l|c|c|}
		\hline
		\textbf{Method} & \textbf{PSNR} & \textbf{SSIM} \\ \hline
		PNLS \cite{xu2021} & 33.8614 & 0.9623 \\ \hline
		$\ell_0$ Smoothing \cite{xu2011image} & 32.2160 & 0.9517 \\ \hline  
		Pottslab \cite{storath2014fast} & 33.6690 & 0.9594 \\ \hline
        Region Fusion \cite{nguyen2015fast} & 33.1948	  &  0.9586\\ \hline
		GSF \cite{liu2021generalized} & 33.8940 & \textcolor{blue}{\textbf{0.9649}} \\ \hline
		RTV \cite{xu2012structure}& 33.5896 & 0.9592 \\ \hline
		Semi-Global WLS \cite{liu2017semi} & 33.4778 & 0.9576 \\ \hline
		Semi-Sparsity Filter \cite{huang2023semi} & 32.9694 & 0.9528 \\ \hline
		JESS-Net \cite{feng2022}  & \textcolor{blue}{\textbf{34.4435}} & \textbf{0.9664} \\ \hline
		ResNet \cite{zhu2019benchmark} & 29.0895 & 0.9326 \\ \hline
		VDCNN \cite{zhu2019benchmark} & 29.8640 & 0.9353 \\ \hline
		Deep WLS \cite{yang2024weighted} & 33.3790 & 0.9531 \\ \hline
         DIP \cite{ulyanov2018deep}& 29.3037 & 0.9033 \\ \hline
        DIP-TV \cite{liu2019image} & 33.8018 & 0.9605 \\ \hline
        Region Fusion DIP-$\ell_0$  (Proposed) & \textbf{34.9755} & 0.9641 \\ \hline
	\end{tabular}

	\label{tab:img_smooth_result}
\end{table}

\captionsetup{indention=0pt,justification=centering}
\begin{figure}[t!]
	\centering
	\begin{subfigure}[t]{0.2\textwidth}
		\centering
		\includegraphics[scale=0.2]{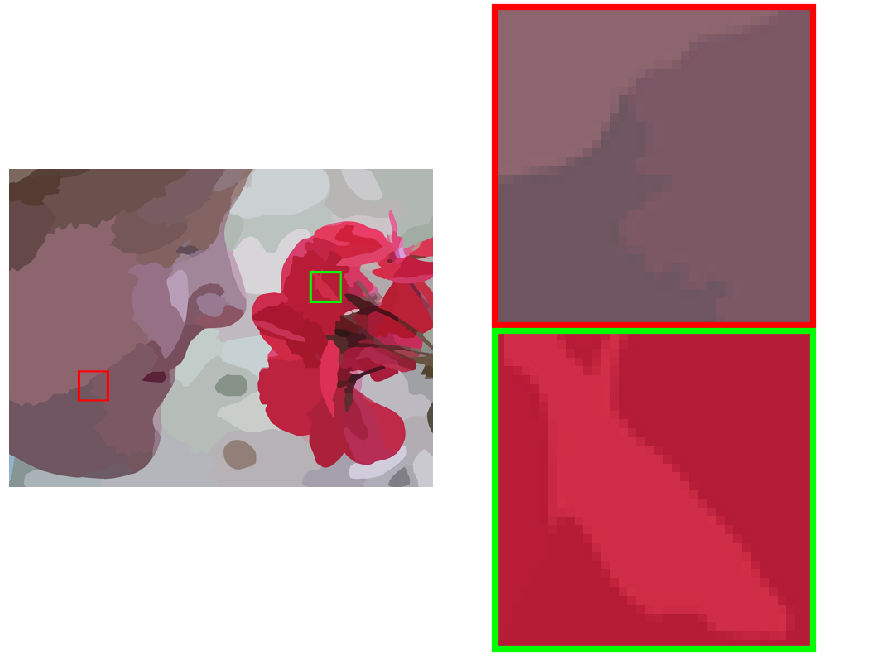}
		\caption{Original}
	\end{subfigure}%
	\begin{subfigure}[t]{0.2\textwidth}
		\centering
		\includegraphics[scale=0.2]{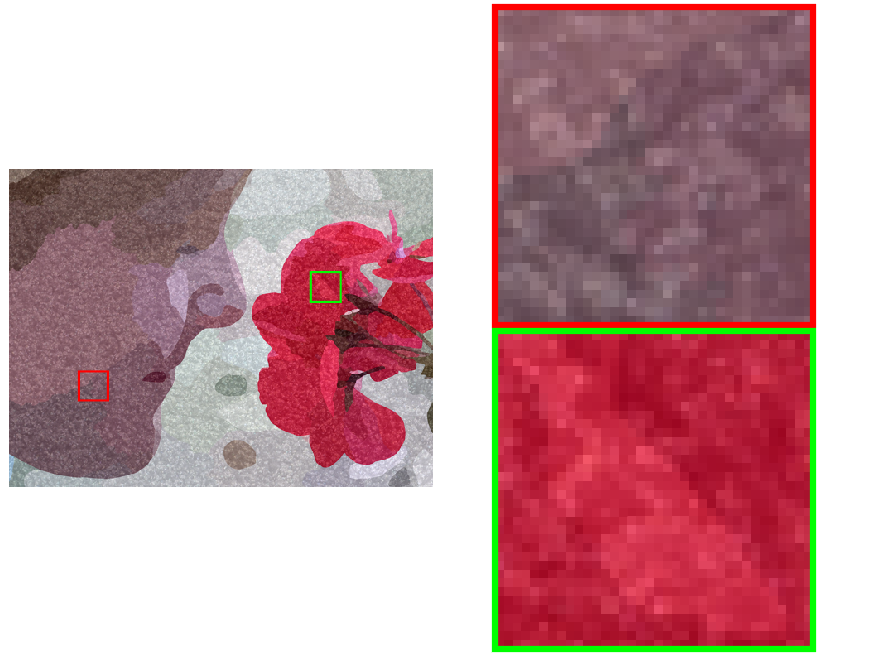}
		\caption{Corrupted}
	\end{subfigure}
	\begin{subfigure}[t]{0.2\textwidth}
		\centering
		\includegraphics[scale=0.2]{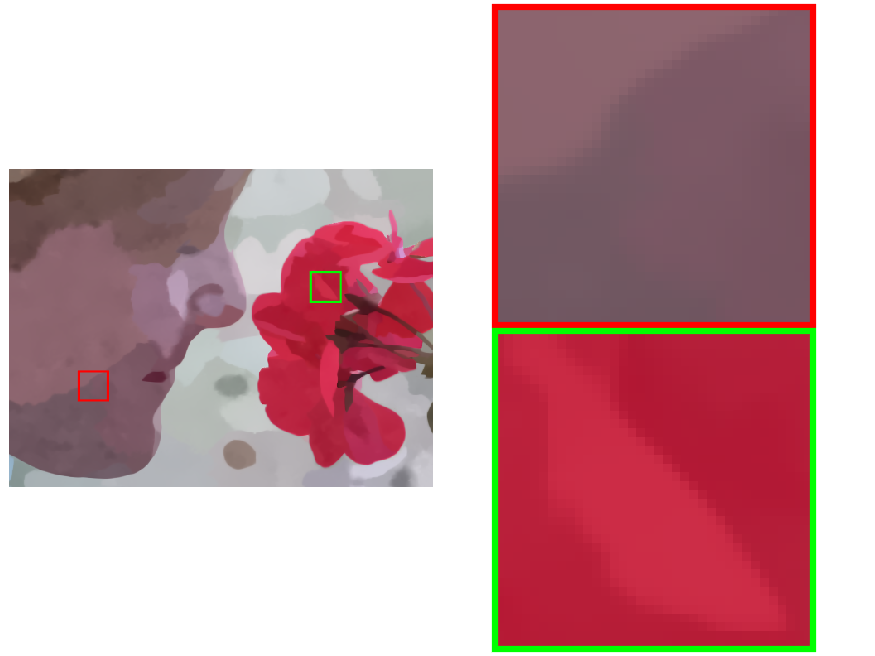}
		\caption{PNLS \cite{xu2021} \\
			34.6099/0.9811}
	\end{subfigure}%
	\begin{subfigure}[t]{0.2\textwidth}
		\centering
		\includegraphics[scale=0.2]{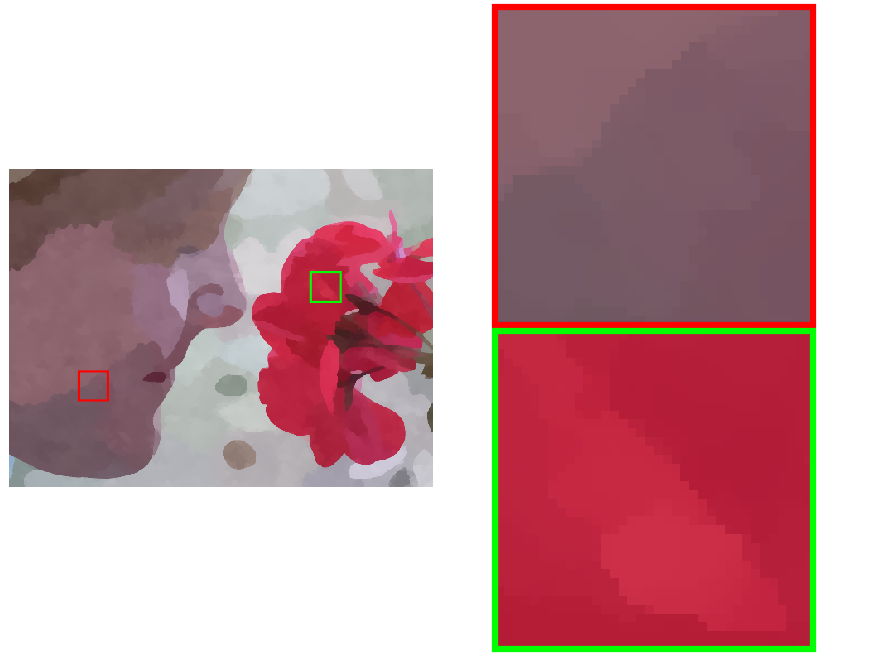}
		\caption{GSF \cite{liu2021generalized} \\
			34.8130/0.9850}
	\end{subfigure}
	\begin{subfigure}[t]{0.2\textwidth}
		\centering
		\includegraphics[scale=0.2]{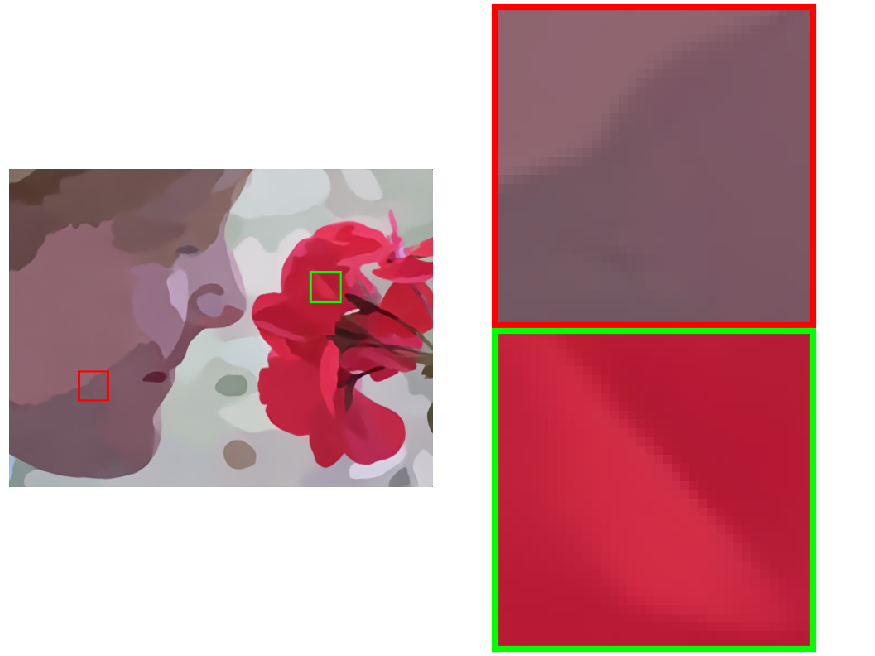}
		\caption{JESS-Net \cite{feng2022}\\
			35.3005/\textbf{0.9859}}
	\end{subfigure}%
		\begin{subfigure}[t]{0.2\textwidth}
		\centering
		\includegraphics[scale=0.2]{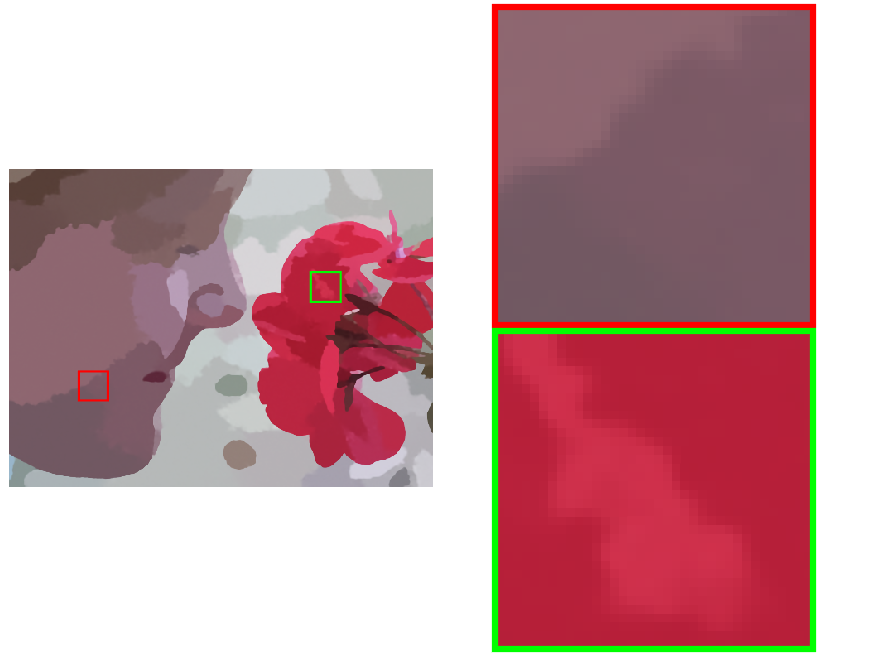}
		\caption{Region Fusion DIP-$\ell_0$ \\
			\textbf{36.7624}/0.9827}
	\end{subfigure}\captionsetup{format=default,indention=0pt,justification=raggedright}
	\caption{Example of image smoothing results on NKS dataset of the top four methods with highest average PSNRs. PSNR/SSIM are shown. \textbf{Bold} denotes best result. }
	\label{fig:smooth_nks_example}
    \vspace{-0.5cm}
\end{figure}

The methods are evaluated on the NKS dataset \cite{xu2021} of 200 images blended with natural texture structures. We randomly split the dataset into 80 images as the validation set for parameter tuning and the remaining 120 images as the test set for evaluation. 
The optimal parameters for Algorithm \ref{alg:ADMM_DIP} are  $\lambda = 0.025, \beta = 2.25, T=100$, and $\alpha = 0.001$.

The results are shown in Table \ref{tab:img_smooth_result}. Region Fusion DIP-$\ell_0$ attains the best PSNR, but its SSIM is slightly behind only JESS-Net and GSF. Although it solves a similar optimization problem as $\ell_0$ smoothing, Pottslab, and Region Fusion, it significantly outperforms them because the DNN is able to output a smoothed image that better approximates  the input image by the fidelity term in \eqref{eq:dip_l0}. DIP-$\ell_0$ along with JESS-Net demonstrate that DNNs can more powerfully extract the structural information of an image than a local or global filter.  However, if they overfit on a training set or become similar to a non-deep filter, their capability can become limited, as exemplified by Resnet, VDCNN, and Deep WLS. Overall, our proposed method can outperform most filters, even deep filters.  

\begin{figure}[t!]
	\centering
	\begin{subfigure}[t]{0.085\textwidth}
		\centering
		\includegraphics[height=0.4in]{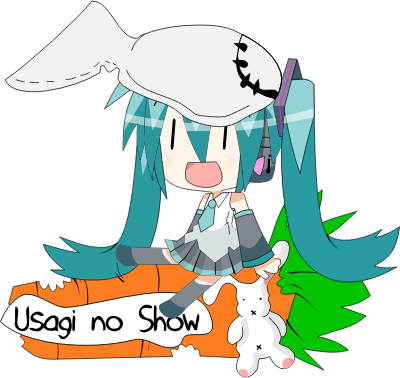}
		\caption{}
		\label{fig:miku}
	\end{subfigure}%
	\begin{subfigure}[t]{0.085\textwidth}
		\centering
		\includegraphics[height=0.4in]{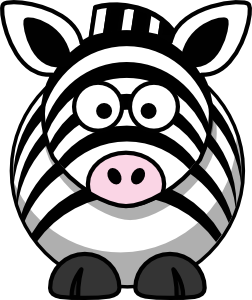}
		\caption{}
	\end{subfigure}%
	\begin{subfigure}[t]{0.085\textwidth}
		\centering
		\includegraphics[height=0.4in]{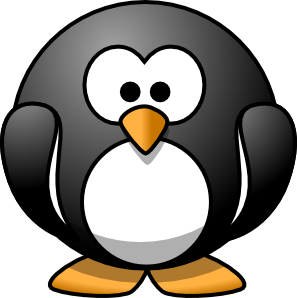}
		\caption{}
	\end{subfigure}%
	\begin{subfigure}[t]{0.085\textwidth}
		\centering
		\includegraphics[height=0.4in]{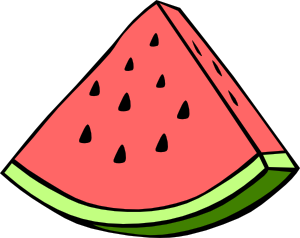}
		\caption{}
	\end{subfigure}%
	\begin{subfigure}[t]{0.085\textwidth}
		\centering
		\includegraphics[height=0.4in]{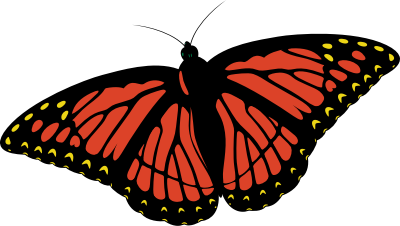}
		\caption{}
	\end{subfigure}
	\begin{subfigure}[t]{0.085\textwidth}
		\centering
		\includegraphics[height=0.4in]{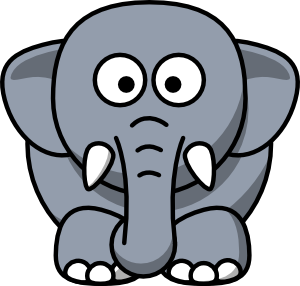}
		\caption{}
	\end{subfigure}%
	\begin{subfigure}[t]{0.085\textwidth}
		\centering
		\includegraphics[height=0.4in]{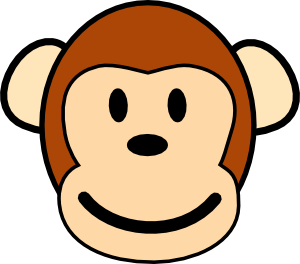}
		\caption{}
	\end{subfigure}%
	\begin{subfigure}[t]{0.085\textwidth}
		\centering
		\includegraphics[height=0.4in]{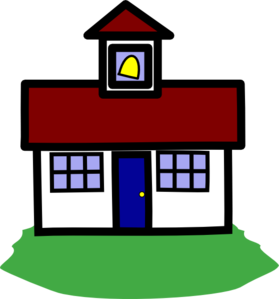}
		\caption{}
	\end{subfigure}%
	\begin{subfigure}[t]{0.085\textwidth}
		\centering
		\includegraphics[height=0.4in]{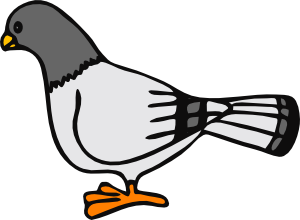}
		\caption{}
	\end{subfigure}%
	\begin{subfigure}[t]{0.085\textwidth}
		\centering
		\includegraphics[height=0.4in]{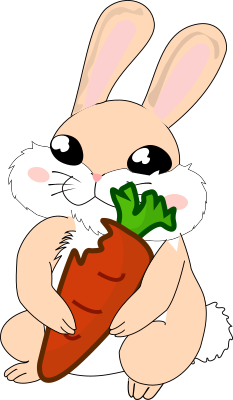}
		\caption{}
        \label{fig:bunny}
	\end{subfigure}
	\caption{10 clip art images for the JPEG compression artifact removal experiments.}
    \vspace{-0.5cm}
	\label{fig:20_images_jpeg_art_remove}
\end{figure}

Fig. \ref{fig:smooth_nks_example} presents visual results of the top four algorithms applied to a test set image. Region fusion DIP-$\ell_0$ has the highest PSNR while JESS-Net has the highest SSIM. In the overall image results, Region Fusion DIP-$\ell_0$ and JESS-Net have smoother regions in the face than the other methods. Zooming in on the face, results are similar between the four methods. Zooming in on the flower, the ``island" region appears sharper in the proposed method than in JESS-Net. However, the edges are more defined in GSF's result.

\subsection{JPEG Artifact Removal in Clip Art}
\begin{table*}[!ht]
        \caption{PSNR/SSIM metrics on JPEG compression artifact removal with a compression quality factor of 10\%.  \textbf{Black bold} is best; \textcolor{blue}{\textbf{blue bold}} is second best.}
    \centering
    \resizebox{\textwidth}{!}{\begin{tabular}{|l|c|c|c|c|c|c|c|c|c|c|c|}
    \hline
        Method & (a) & (b) & (c) & (d) & (e) & (f) & (g) & (h) & (i) & (j)& Avg. \\ \hline
        PNLS \cite{xu2021} & 22.028/0.9156 & 17.023/0.7486 & 22.169/0.8649 & 27.298/0.9653 & 24.570/0.8428 & 23.582/0.9029 & 28.404/0.9431 & 24.615/0.9169 & 21.004/0.8870 & 22.995/0.9274 & 23.369/0.8914 \\ \hline
        $\ell_0$ Smoothing \cite{xu2011image} & 26.412/0.9619 & 27.118/0.9005 & 30.116/0.9080 & 28.392/0.9755 & 25.174/0.8482 & 29.561/0.9546 & 29.773/0.9501 & 28.571/0.9320 & 29.539/0.9551 & 26.140/0.9612 & 28.080/0.9347 \\ \hline
         Pottslab \cite{storath2014fast} & 26.315/0.9632 & 27.422/0.9101 & 30.283/0.9067 & 28.732/0.9756 & 25.177/\textcolor{blue}{\textbf{0.8595}} & 29.736/0.9614 & 29.975/0.9538 & 28.547/0.9369 & 29.917/0.9609 & 26.410/0.9663 & 28.251/0.9394 \\ \hline
        Region Fusion \cite{nguyen2015fast} & 26.311/0.9623 & 27.179/0.9062 & 30.069/0.9050 & 28.228/0.9734 & 24.930/0.8587 & 29.365/0.9585 & 29.679/0.9525 & 28.669/0.9384 & 29.667/0.9629 & 26.255/0.9651 & 28.035/0.9383 \\ \hline
        GSF \cite{liu2021generalized} & 26.093/0.9601 & 26.656/\textbf{0.9480} & 30.540/\textbf{0.9477} & 29.190/0.9762 & \textcolor{blue}{\textbf{25.581}}/\textbf{0.8691} & 29.255/\textcolor{blue}{\textbf{0.9653}} & 29.715/\textbf{0.9591} & 28.664/\textbf{0.9411} & 29.734/0.9730 & 25.975/0.9659 & 28.140/\textbf{0.9506} \\ \hline
        RTV \cite{xu2012structure} & 26.215/0.9580 & 27.552/0.9123 & 30.441/0.9162 & \textbf{29.239}/\textbf{0.9772} & 25.514/0.8442 & 30.205/0.9652 & \textcolor{blue}{\textbf{30.574}}/0.9540 & \textcolor{blue}{\textbf{29.159}}/0.9342 & 30.478/0.9631 & 25.959/0.9611 & 28.534/0.9386 \\ \hline
        Semi-Global WLS \cite{liu2017semi} & 26.556/0.9625 & 27.327/0.9029 & 31.024/0.9249 & \textcolor{blue}{\textbf{29.235}}/\textcolor{blue}{\textbf{0.9768}} & \textbf{25.782}/0.8542 & 29.741/0.9588 & 30.376/0.9532 & 28.912/0.9355 & 30.318/0.9692 & 26.452/0.9649 & 28.572/0.9403 \\ \hline
        Semi-Sparsity Filter \cite{huang2023semi} & 26.294/0.9595 & 27.032/0.9091 & 30.198/0.9242 & 27.915/0.9705 & 24.830/0.8300 & 29.093/0.9485 & 29.008/0.9428 & 27.999/0.9264 & 29.441/0.9700 & 25.923/0.9587 & 27.773/0.9340 \\ \hline
        JESS-Net \cite{feng2022} & 25.834/0.9571 & 27.668/\textcolor{blue}{\textbf{0.9406}} & 30.793/0.9125 & 28.134/0.9712 & 23.208/0.8125 & 29.428/0.9479 & 29.789/\textcolor{blue}{\textbf{0.9564}} & 27.618/0.9263 & 29.328/0.9606 & 26.034/0.9673 & 27.783/0.9352 \\ \hline
        ResNet \cite{zhu2019benchmark} & 26.291/0.9610 & 26.910/0.8544 & 28.354/0.8778 & 27.641/0.9598 & 24.780/0.7920 & 29.515/0.9405 & 29.109/0.9268 & 28.433/0.9033 & 30.251/\textbf{0.9824} & 25.944/0.9610 & 27.723/0.9159 \\ \hline
        VDCNN \cite{zhu2019benchmark} & 26.116/0.9609 & 27.049/0.8901 & 29.768/0.9048 & 27.661/0.9657 & 24.907/0.8166 & 29.513/0.9500 & 29.250/0.9401 & 28.525/0.9218 & 30.095/0.9758 & 25.988/0.9629 & 27.887/0.9289 \\ \hline
        Deep WLS \cite{yang2024weighted} & 26.265/0.9595 & 27.053/0.8945 & 30.485/0.9250 & 27.890/0.9628 & 24.948/0.8282 & 29.316/0.9485 & 29.158/0.9384 & 28.564/0.9237 & 30.170/0.9761 & 26.132/0.9622 & 27.998/0.9319 \\ \hline
        DIP \cite{ulyanov2018deep} & \textcolor{blue}{\textbf{26.820}}/0.9621 & \textbf{28.176}/0.9089 & \textcolor{blue}{\textbf{31.597}}/0.9369 & 28.096/0.9675 & 24.968/0.8123 & \textcolor{blue}{\textbf{30.465}}/0.9620 & 30.079/0.9462 & 28.968/0.9263 & \textcolor{blue}{\textbf{30.960}}/0.9761 & \textcolor{blue}{\textbf{26.757}}/\textcolor{blue}{\textbf{0.9684}} & \textcolor{blue}{\textbf{28.688}}/0.9367 \\ \hline
        DIP-TV \cite{liu2019image} & 26.807/\textcolor{blue}{\textbf{0.9646}} & 27.491/0.9058 & 31.344/0.9397 & 28.028/0.9662 & 24.879/0.8257 & 29.981/0.9592 & 29.778/0.9450 & 28.868/0.9275 & 30.552/\textcolor{blue}{\textbf{0.9781}} & 26.684/0.9681 & 28.441/0.9380 \\ \hline
        Region Fusion DIP-$\ell_0$ & \textbf{27.373}/\textbf{0.9703} & \textcolor{blue}{\textbf{27.996}}/0.9138 & \textbf{32.494}/\textcolor{blue}{\textbf{0.9418}} & 29.149/0.9763 & 25.603/0.8580 & \textbf{30.911}/\textbf{0.9673} & \textbf{30.679}/0.9542 & \textbf{29.453}/\textcolor{blue}{\textbf{0.9398}} & \textbf{31.023}/0.9724 & \textbf{27.020}/\textbf{0.9703} & \textbf{29.170}/\textcolor{blue}{\textbf{0.9464}} \\ \hline
    \end{tabular}}

            \label{tab:jpeq_result}
    \vspace{-0.5cm}
\end{table*}
\begin{figure}[t!]
    \centering
    \begin{subfigure}[t]{0.225\textwidth}
        \centering
        \includegraphics[scale=0.20]{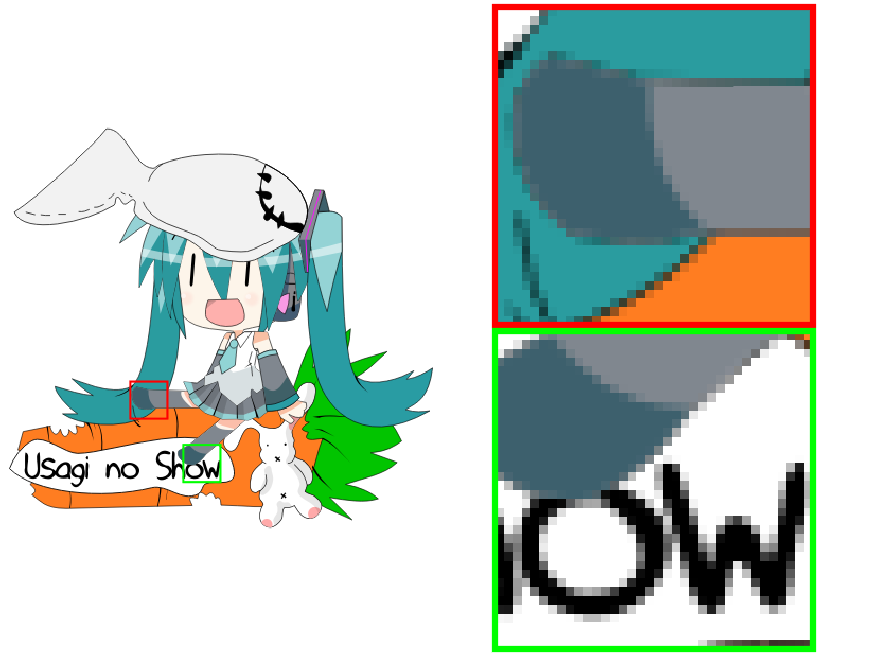}
        \caption{Original}
    \end{subfigure}%
    \begin{subfigure}[t]{0.225\textwidth}
        \centering
        \includegraphics[scale=0.20]{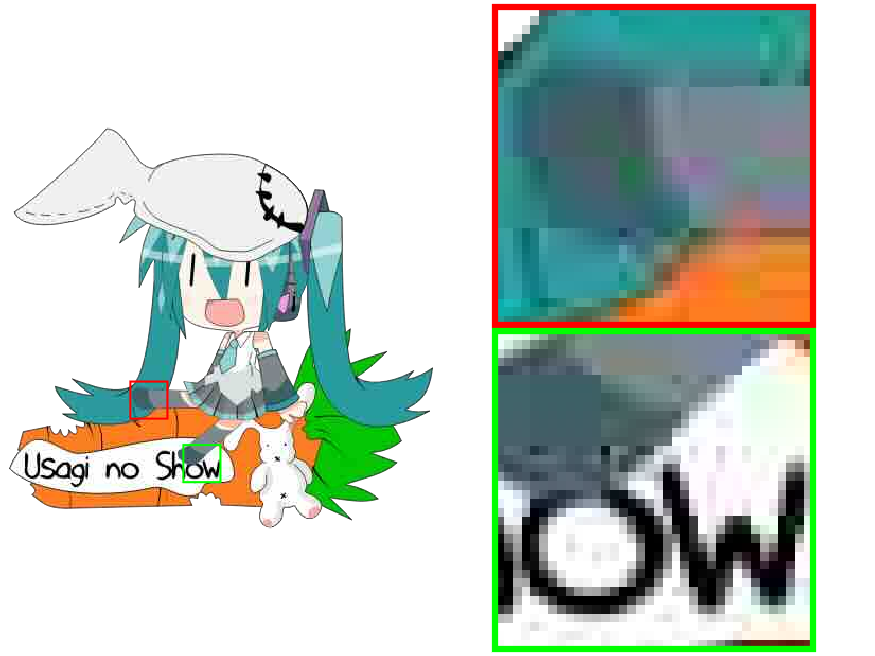}
        \caption{JPEG 10\%}
        \label{fig:jpeg_miku}
    \end{subfigure}
        \begin{subfigure}[t]{0.225\textwidth}
        \centering
        \includegraphics[scale=0.20]{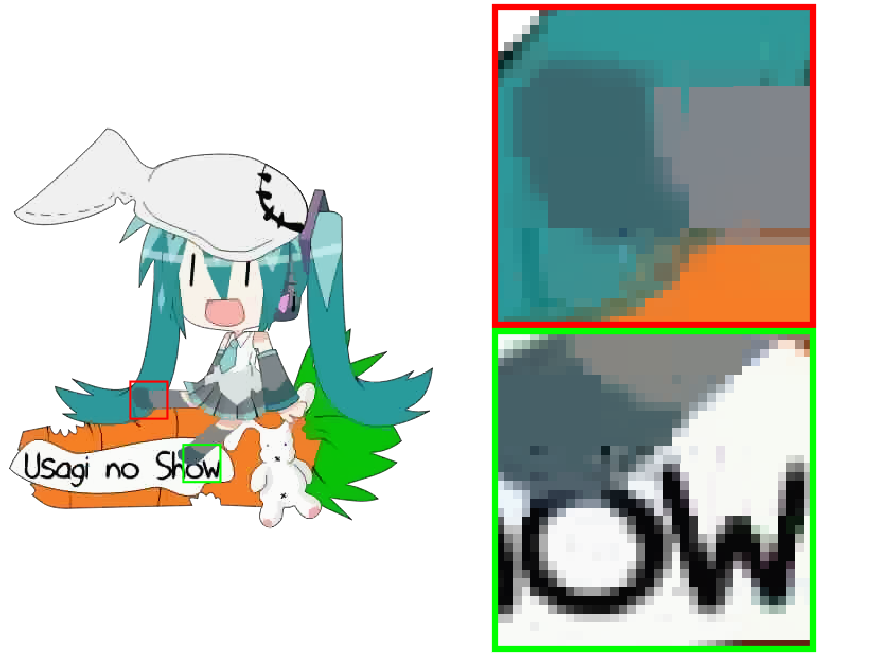}
        \caption{Semi-Global WLS \cite{liu2017semi}}
    \end{subfigure}%
                \begin{subfigure}[t]{0.225\textwidth}
        \centering
        \includegraphics[scale=0.20]{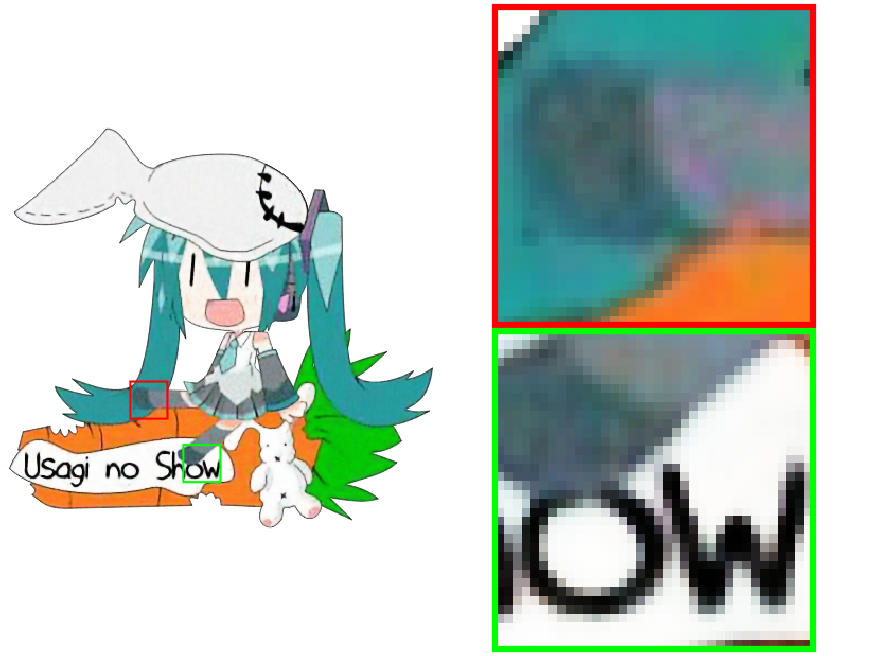}
        \caption{DIP \cite{ulyanov2018deep}}
    \end{subfigure}
        \begin{subfigure}[t]{0.225\textwidth}
        \centering
        \includegraphics[scale=0.20]{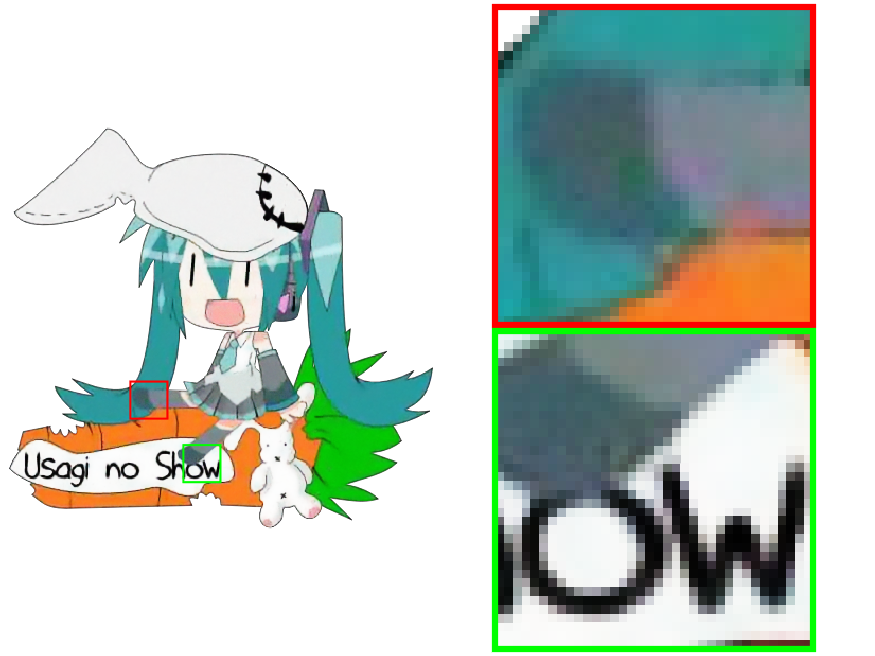}
        \caption{DIP-TV \cite{liu2019image}}
    \end{subfigure}%
    \begin{subfigure}[t]{0.225\textwidth}
        \centering
        \includegraphics[scale=0.20]{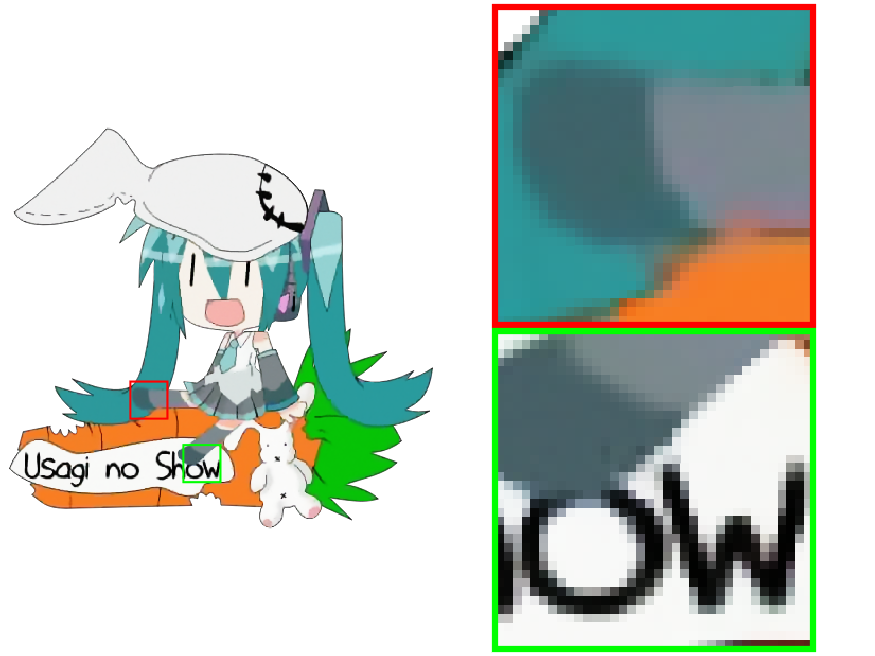}
        \caption{Region Fusion DIP-$\ell_0$}
    \end{subfigure}%
    \caption{Top four results on JPEG compression artifact removal on Fig. \ref{fig:miku} with compression quality factor of 10\%. }
    \label{fig:jpeg_art_move_miku_result}
    \vspace{-0.5cm}
\end{figure}
We compare the performances of the smoothing algorithms in removing JPEG artifacts, such as mosquito noises and blur. Shown in Fig. \ref{fig:20_images_jpeg_art_remove}, our evaluation dataset consists of 10 clip arts collected online. The clip arts are compressed with a quality factor of 10\%. For Algorithm \ref{alg:ADMM_DIP}, the best parameter combination is
$\lambda = 0.025, \beta = 2.0, T=100$, and $\alpha = 0.001$.

Table \ref{tab:jpeq_result} reports the PSNRs/SSIMs for each clip art image. On average, Region Fusion DIP-$\ell_0$ attains the best average PSNR and the second best average SSIM. It is the best method for 7 images by PSNR and the best or second best method for 5 images by SSIM. Some other algorithms are only the best for one image by PSNR, while GSF is the only other method that ranks the best for 5 images by SSIM. Thus, our proposed method is consistent in effectively removing JPEG artifacts. 

Fig. \ref{fig:jpeg_art_move_miku_result} presents some results of the JPEG-compressed Fig. \ref{fig:miku} smoothed by the top four methods by average PSNR. Region Fusion DIP-$\ell_0$ restore most of the edges, such as in the leg shown in the top zoom-in image, better than DIP and DIP-TV, which has some discoloration and blur. Moreover, it removes most of the dark spots in the bottom zoom-in image of Fig. \ref{fig:jpeg_miku}. In the bottom zoom-in image, although semi-global WLS removes the block artifacts around the text, some still remain in the ``foot" region. Our proposed method is the most effective in restoring Fig. \ref{fig:jpeg_miku}.

\begin{figure}[t!]
	\centering
	\begin{subfigure}[t]{0.18\textwidth}
		\centering
		\includegraphics[height=1in]{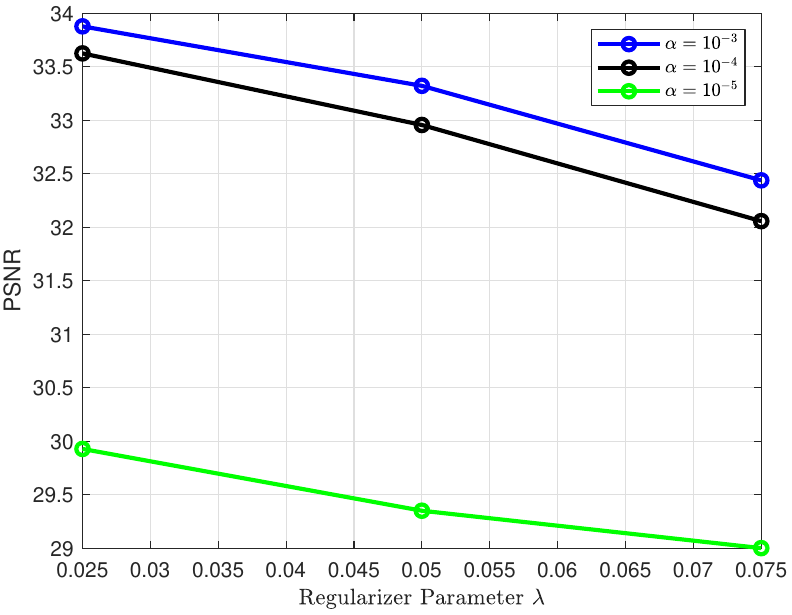}
		\caption{Changes in PSNR with respect to $\lambda$ with $\beta = 2.25, T=100$.}
		\label{fig:lambda_analysis}
	\end{subfigure}%
	\begin{subfigure}[t]{0.18\textwidth}
		\centering
		\includegraphics[height=1in]{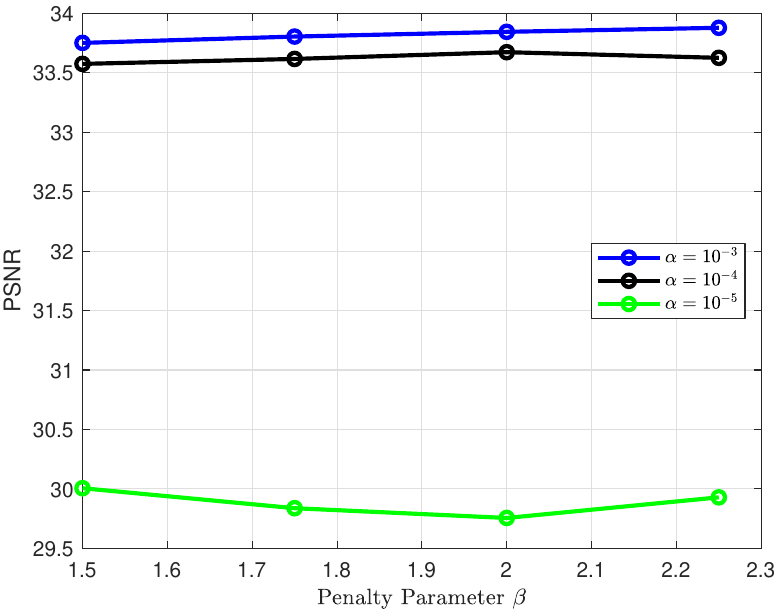}
		\caption{Changes in PSNR with respect to $\beta$ with $\lambda =  0.025, T=100$.}
	\end{subfigure}
	\begin{subfigure}[t]{0.18\textwidth}
		\centering
		\includegraphics[height=1in]{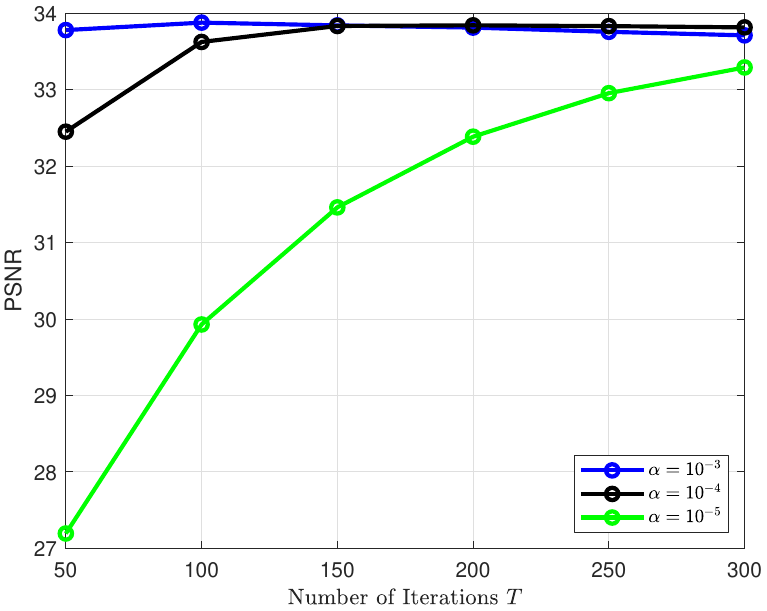}
		\caption{Changes in PSNR with respect to $T$ with $\lambda = 0.025, \beta  = 2.25$.}
	\end{subfigure}
	\caption{Sensitivity analysis on the model parameters $\lambda, \beta,$  and $T$ on the average PSNR of 20 images of the NKS dataset.}
	\label{fig:param_analysis}
    \vspace{-0.75cm}
\end{figure}

\subsection{Parameter Analysis}
We analyze how the learning rate $
\alpha$, regularizer parameter $\lambda$, penalty parameter $\beta$, and the number of iterations $T$ impact the performance of Algorithm \ref{alg:ADMM_DIP}, in particular PSNR since DIP-$\ell_0$ performs best in that metric. To perform the analysis, we apply Algorithm \ref{alg:ADMM_DIP} to 20 images of NKS dataset. We examine the learning rate $\alpha \in \{10^{-3}, 10^{-4}, 10^{-5}\}$ while we vary one parameter at a time. We fix $\lambda = 0.025, \beta = 2.25$, and $T=100$ when they are not being changed. Figure \ref{fig:param_analysis} shows the impact of the parameters on the average PSNR of the results. In general, we observe that the learning rate $\alpha = 10^{-3}$ yields the best results across the different combinations of parameters. As $\lambda$ increases, the $\ell_0$ gradient regularizer tends to remove more edges or details than necessary, thereby decreasing the average PSNR. Different values of the penalty parameter $\beta$ result in similar PSNRs across different learning rates. Lastly, the average PSNR increases as $T$ increases for learning rates $\alpha =10^{-4}, 10^{-5}$. However, it decreases after 100 iterations for $\alpha = 10^{-3}$, likely resulting in overfitting. Therefore, a decently large learning rate, small value for $\lambda$, and $T=100$ are recommended for Algorithm \ref{alg:ADMM_DIP}.
\subsection{Limitation}
Although our proposed DIP-$\ell_0$ demonstrates remarkable performance in our experiments, it can be computationally slow because it needs to train the DNN weights for each image. Table \ref{tab:dip} reports that running DIP-$\ell_0$ for $T=100$ iterations can take a few minutes, conveying that the bulk of the computation is due to the forward and backward passes in DNNs.
\begin{table}[!t]
    \centering
        \caption{Computational times in seconds on two images. Pottslab and Region Fusion are performed on CPU. DIP-$\ell_0$ is ran on  NVIDIA GeForce RTX
2080 for $T=100$ iterations. }
    \scriptsize
    \begin{tabular}{|l|c|c|}
    \hline
        Figure (size)& Region Fusion & \makecell{Region Fusion\\ DIP-$\ell_0$} \\ \hline
        Fig. \ref{fig:miku} (400x378) &  0.25 & 215.97 \\ \hline
        Fig. \ref{fig:bunny} (233x400) & 0.15 & 146.72 \\ \hline
    \end{tabular}
\vspace{-0.2cm}
\label{tab:dip}
\end{table}
\section{Conclusion}
In this paper, we propose DIP-$\ell_0$, a DIP framework for image smoothing that incorporates the $\ell_0$ gradient regularizer. The $\ell_0$ gradient regularizer encourages the output of DIP-$\ell_0$ to be piecewise constant so that only the strong edges of the original input image remains, leading to a smoothed image result. Because the loss function of DIP-$\ell_0$ contains the nonconvex, nonsmooth $\ell_0$ ``norm", we develop an ADMM algorithm to properly minimize it. Our experiments demonstrate that DIP-$\ell_0$ performs well in edge-preserving image smoothing and JPEG artifact removal. Although DIP-$\ell_0$ is an unsupervised deep filter, it outperforms supervised deep filters that may have overfitted on the training dataset. In the future, we will examine DIP-$\ell_0$ in other applications, such as image enhancement and pencil sketching, and improve its computational efficiency, such as using deep random projector \cite{li2023deep}.

\bibliographystyle{IEEEbib}
\bibliography{strings,refs}

\end{document}